\documentclass{article}

\usepackage[nonatbib,final]{neurips_2021}

\usepackage[square,sort,comma,numbers]{natbib}

\usepackage[utf8]{inputenc} 
\usepackage[T1]{fontenc}    
\usepackage{hyperref}       
\hypersetup{
    colorlinks=true,
    linkcolor=blue,
    filecolor=magenta,      
    urlcolor=blue,
}
\usepackage{url}            
\usepackage{booktabs}       
\usepackage{amsfonts}       
\usepackage{nicefrac}       
\usepackage{microtype}      

\usepackage{algorithm}
\usepackage{algorithmic}
\usepackage{wrapfig}
\usepackage{enumitem}
\usepackage{subfigure}

\usepackage{amssymb,amsmath}
\usepackage{mathtools}

\usepackage{lipsum} 

\usepackage[table,xcdraw]{xcolor} 
\usepackage{multirow}
\usepackage[normalem]{ulem} 
\usepackage{caption}
\usepackage{comment}

\usepackage{footmisc}
\DefineFNsymbols{mySymbols}{{\ensuremath\dagger}{\ensuremath\ddagger}\S\P
   *{**}{\ensuremath{\dagger\dagger}}{\ensuremath{\ddagger\ddagger}}}
\setfnsymbol{mySymbols}

\title{The Sensory Neuron as a Transformer: Permutation-Invariant Neural Networks for Reinforcement Learning}

\author{
  Yujin Tang\thanks{Equal Contribution} \\
  Google Brain\\
  \texttt{yujintang@google.com} \\
  \And
  David Ha\textsuperscript{\ensuremath\dagger}\\
  Google Brain\\
  \texttt{hadavid@google.com} \\
}

\begin{document}
\maketitle
\vskip -0.195in 
\begin{abstract}
\vskip -0.1in 
In complex systems, we often observe complex global behavior emerge from a collection of agents interacting with each other in their environment, with each individual agent acting only on locally available information, without knowing the full picture. Such systems have inspired development of artificial intelligence algorithms in areas such as swarm optimization and cellular automata. Motivated by the emergence of collective behavior from complex cellular systems, we build systems that feed each sensory input from the environment into distinct, but identical neural networks, each with no fixed relationship with one another. We show that these sensory networks can be trained to integrate information received locally, and through communication via an attention mechanism, can collectively produce a globally coherent policy. Moreover, the system can still perform its task even if the ordering of its inputs is randomly permuted several times during an episode. These permutation invariant systems also display useful robustness and generalization properties that are broadly applicable. Interactive demo and videos of our results: \url{https://attentionneuron.github.io/}
\end{abstract}
\section{Introduction}
\vskip -0.1in 

Sensory substitution refers to the brain's ability to use one sensory modality (e.g., touch) to supply environmental information normally gathered by another sense (e.g., vision). Numerous studies have demonstrated that humans can adapt to changes in sensory inputs, even when they are fed into the \textit{wrong} channels~\cite{bach1969vision,bach2003sensory,sandlin2019backwards,eagleman2020livewired}.
But difficult adaptations--such as learning to ``see'' by interpreting visual information emitted from a grid of electrodes placed on one's tongue~\cite{bach2003sensory}, or learning to ride a ``backwards'' bicycle~\cite{sandlin2019backwards}--require months of training to attain mastery.
Can we do better, and create artificial systems that can rapidly adapt to sensory substitutions, without the need to be retrained?

\begin{figure}[!htb]   
\vskip -0.10in 
\centering        
\includegraphics[width=1\textwidth]{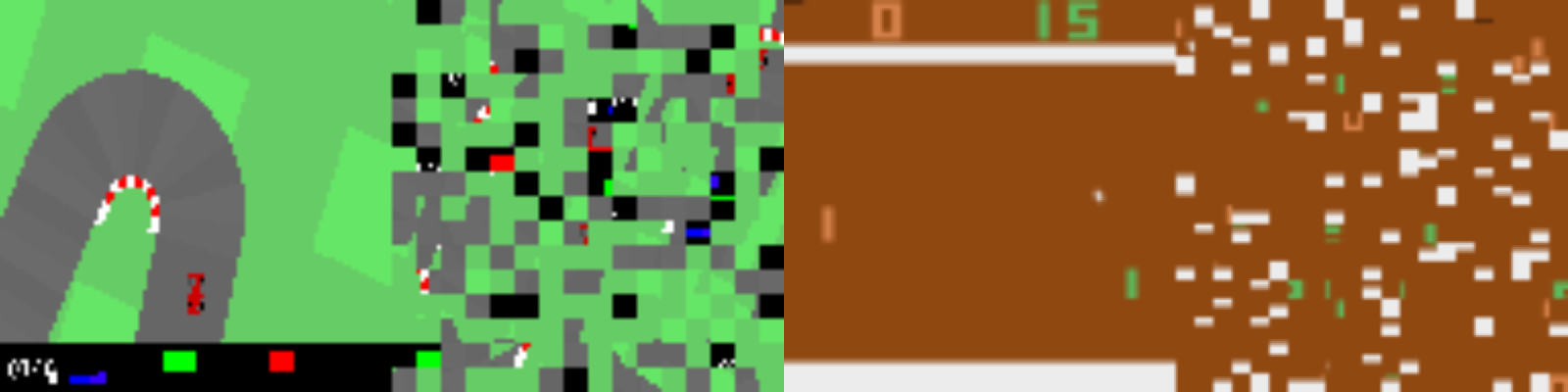}     
\vskip -0.00in 
\caption{
\textit{Comparison of visual input intended for the game player, and what our system receives.}
\newline
We partition the visual input from CarRacing (Left) and Atari Pong (right) into a 2D grid of small patches, and randomly permute their ordering. Each sensory neuron in the system receives a stream of visual input at a particular permuted patch location, and through coordination, must complete the task at hand, even if the visual ordering is randomly permuted again several times during an episode.
}
\label{fig:cover_diagram} 
\vskip -0.03in 
\end{figure}


Modern deep learning systems are generally unable to adapt to a sudden reordering of sensory inputs, unless the model is retrained, or if the user manually corrects the ordering of the inputs for the model. However, techniques from continual meta-learning, such as adaptive weights~\cite{schmidhuber1992learning,ba2016using,ha2016hypernetworks}, Hebbian-learning~\cite{miconi2018differentiable,miconi2020backpropamine,najarro2020meta}, and model-based~\cite{deisenroth2011pilco,amos2018differentiable,ha2018worldmodels,hafner2018planet} approaches can help the model adapt to such changes, and remain a promising active area of research.

In this work, we investigate agents that are explicitly designed to deal with sudden random reordering of their sensory inputs while performing a task. Motivated by recent developments in self-organizing neural networks~\cite{fortuin2018som,mordvintsev2020growing,randazzo2020selfclassifying} related to cellular automata~\cite{neumann1966theory,codd2014cellular,conway1970game,wolfram1984cellular,chopard1998cellular}, in our experiments, we feed each sensory input (which could be an individual state from a continuous control environment, or a patch of pixels from a visual environment) into an individual neural network module that integrates information from only this particular sensory input channel over time. While receiving information locally, each of these individual sensory neural network modules also continually broadcasts an output message. Inspired by the Set Transformer~\cite{vaswani2017,set2019} architecture, an attention mechanism combines these messages to form a global latent code which is then converted into the agent's action space. The attention mechanism can be viewed as a form of adaptive weights of a neural network, and in this context, allows for an arbitrary number of sensory inputs that can be processed in any random order.

In our experiments, we find that each individual sensory neural network module, despite receiving only localized information, can still collectively produce a globally coherent policy, and that such a system can be trained to perform tasks in several popular reinforcement learning (RL) environments. Furthermore, our system can utilize a varying number of sensory input channels in any randomly permuted order, even when the order is shuffled again several times during an episode.

Permutation invariant systems have several advantages over traditional fixed-input systems.
We find that encouraging a system to learn a coherent representation of a permutation invariant observation space leads to policies that are more robust and generalizes better to unseen situations.
We show that, without additional training, our system continues to function even when we inject additional input channels containing noise or redundant information.
In visual environments, we show that our system can be trained to perform a task even if it is given only a small fraction of randomly chosen patches from the screen, and at test time, if given more patches, the system can take advantage of the additional information to perform better.
We also demonstrate that our system can generalize to visual environments with novel background images, despite training on a single fixed background.
Lastly, to make training more practical, we propose a behavioral cloning scheme to convert policies trained with existing methods into a permutation invariant policy with desirable properties.

\section{Related Work}
\vskip -0.1in 

\textbf{Self-organization} is a process where some form of global order emerges from local interactions between parts of an initially disordered system.
It is also a property observed in cellular automata (CA)~\cite{neumann1966theory,codd2014cellular,conway1970game}, which are mathematical systems consisting of a grid of cells that perform computation by having each cell communicate with its immediate neighbors and performing a local computation to update its internal state.
Such local interactions are useful in modeling complex systems~\cite{wolfram1984cellular} and have been applied to model non-linear dynamics in various fields~\cite{chopard1998cellular}. Cellular Neural Networks~\cite{chua1988cellular} were first introduced in the 1980s to use neural networks in place of the algorithmic cells in CA systems. They were applied to perform image processing operations with parallel computation. Eventually, the concept of self-organizing neural networks found its way into deep learning in the form of Graph Neural Networks (GNN)~\cite{wu2020comprehensive,sanchezlengeling2021a}.

Using modern deep learning tools, recent work demonstrates that \textit{neural CA}, or self-organized neural networks performing only local computation, can generate (and re-generate) coherent images~\cite{mordvintsev2020growing} and voxel scenes~\cite{zhang2021learning,sudhakaran2021growing}, and even perform image classification~\cite{randazzo2020selfclassifying}. Self-organizing neural network agents have been proposed in the RL domain~\cite{cheney2014unshackling,ohsawa2018neuron,ott2020giving,chang2020decentralized}, with recent work demonstrating that shared local policies at the actuator level~\cite{huang2020}, through communicating with their immediate neighbors, can learn a global coherent policy for continuous control locomotion tasks.
While existing CA-based approaches present a modular, self-organized solution, they are \textit{not} inherently permutation invariant. In our work, we build on neural CA, and enable each cell to communicate beyond its immediate neighbors via an attention mechanism that enables permutation invariance.

\textbf{Meta-learning} recurrent neural networks (RNN)~\cite{hochreiter2001learning,haruno2001mosaic,duan2016rl,wang2016learning} have been proposed to approach the problem of learning the learning rules for a neural network using the reward or error signal, enabling meta-learners to learn to solve problems presented outside of their original training domains. The goals are to enable agents to continually learn from their environments in a single lifetime episode, and to obtain much better data efficiency than conventional learning methods such as stochastic gradient descent (SGD). A meta-learned policy that can adapt the weights of a neural network to its inputs during inference time have been proposed in fast weights~\cite{schmidhuber1992learning,schmidhuber1993self}, associative weights~\cite{ba2016using}, hypernetworks~\cite{ha2016hypernetworks}, and Hebbian-learning~\cite{miconi2018differentiable,miconi2020backpropamine} approaches. Recently works~\cite{sandler2021meta,kirsch2020meta} combine ideas of self-organization with meta-learning RNNs, and have demonstrated that modular meta-learning RNN systems not only can learn to perform SGD-like learning rules, but can also discover more general learning rules that transfer to classification tasks on unseen datasets.

In contrast, the system presented here does not use an error or reward signal to meta-learn or fine-tune its policy. But rather, by using the shared modular building blocks from the meta-learning literature, we focus on learning or converting an existing policy to one that is permutation invariant, and we examine the characteristics such policies exhibit in a zero-shot setting, \textit{without} additional training.

\textbf{Attention} can be viewed as an adaptive weight mechanism that alters the weight connections of a neural network layer based on what the inputs are. Linear \textit{dot-product} attention has first been proposed for meta-learning~\cite{schmidhuber1993reducing}, and versions of linear attention with ${softmax}$ nonlinearity appeared later~\cite{graves2014neural,luong2015effective}, now made popular with Transformer~\cite{vaswani2017}. The adaptive nature of attention provided the Transformer with a high degree of expressiveness, enabling it to learn inductive biases from large datasets and have been incorporated into state-of-the-art methods in natural language processing~\cite{devlin2018bert,brown2020language}, image recognition~\cite{dosovitskiy2020image} and generation~\cite{esser2020taming}, audio and video domains~\cite{girdhar2019video,sun2019learning,jaegle2021perceiver}.

Attention mechanisms have found many uses for RL~\cite{sorokin2015deep,choi2017multi,zambaldi2018deep,mott2019towards,attentionagent2020}. Our work here specifically uses attention to enable communication between arbitrary numbers of modules in an RL agent. While previous work~\cite{velivckovic2017graph,monti2017geometric,zhang2018gaan,yun2019graph,joshi2020transformers,goyal2021recurrent} utilized attention as a communication mechanism between independent neural network modules of a GNN, our work focuses on studying the permutation invariant properties of attention-based communication applied to RL agents. Related work~\cite{liu2020pic} used permutation invariant critics to enhance performance of multi-agent RL. Building \cite{guttenberg2016permutation,zaheer2017deep}, Set Transformers~\cite{set2019} investigated the use of attention explicitly for permutation invariant problems that deal with set-structured data, which have provided the theoretical foundation for our work.

\section{Method}
\label{sec:method}
\vskip -0.13in 

\subsection{Background}
\label{sec:method_background}
\vskip -0.11in 


Our goal is to devise an agent that is permutation invariant (PI) in the action space to the permutations in the input space.
While it is possible to acquire a quasi-PI agent by training with randomly shuffled observations and hope the agent's policy network has enough capacity to memorize all the patterns, we aim for a design that achieves true PI even if the agent is trained with fix-ordered observations. Mathematically, we are looking for a non-trivial function $f(x): \mathcal{R}^n \mapsto \mathcal{R}^m$ such that $f(x[{s}]) = f(x)$ for any $x \in \mathcal{R}^n$, and $s$ is any permutation of the indices $\{1, \cdots, n\}$.
A different but closely related concept is permutation equivariance (PE) which can be described by a function $h(x): \mathcal{R}^n \mapsto \mathcal{R}^n$ such that $h(x[{s}]) = h(x)[s]$. Unlike PI, the dimensions of the input and the output must equal in PE.

Self-attentions can be PE. In its simplest form, self-attention is described as $y = \sigma(QK^{\top})V$ where $Q,K \in \mathcal{R}^{n \times d_q}, V \in \mathcal{R}^{n \times d_v}$ are the Query, Key and Value matrices and $\sigma(\cdot)$ is a non-linear function. In most scenarios, $Q, K, V$ are functions of the input $x \in \mathcal{R}^n$ (e.g. linear transformations), and permuting $x$ therefore is equivalent to permuting the rows in $Q, K, V$ and based on its definition it is straightforward to verify the PE property. Set Transformer~\cite{set2019} cleverly replaced $Q$ with a set of learnable seed vectors, so it is no longer a function of input $x$, thus enabling the output to become PI.
A simple, intuitive explanation of the PI property of self-attention is available in Appendix~\ref{sec:pi_explanation}.

\subsection{Sensory Neurons with Attention}
\label{sec:method_actual}
\vskip -0.11in 

\begin{figure}[ht]   
\vskip -0.05in 
\centering        
\includegraphics[width=1\textwidth]{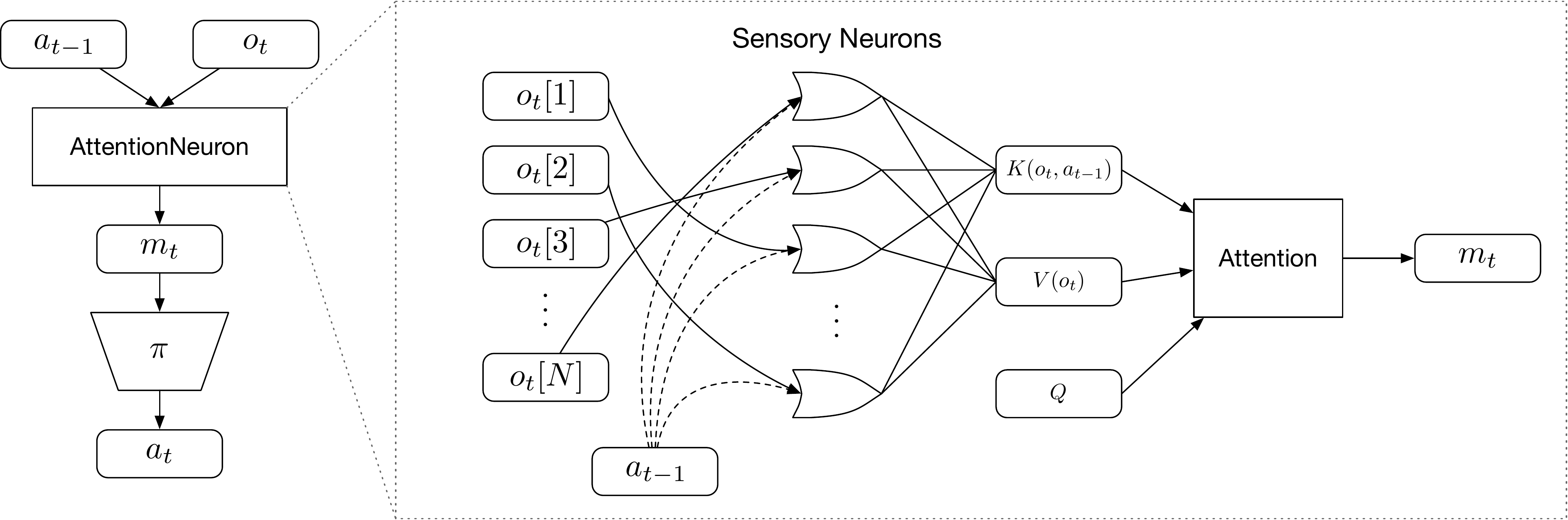}       
\vskip -0.05in 
\caption{
\textit{Overview of Method.}
AttentionNeuron is a standalone layer, in which each sensory neuron only has access to a part of the unordered observations $o_t$. Together with the agent's previous action $a_{t-1}$, each neuron generates messages independently using the shared functions $f_k(o_t[i], a_{t-1})$ and $f_v(o_t[i])$. The attention mechanism summarizes the messages into a global latent code $m_t$.
}
\label{fig:arch} 
\vskip -0.2in 
\end{figure}

To create PI agents, we propose to add an extra layer in front of the agent's policy network $\pi$, which accepts the current observation $o_t$ and the previous action $a_{t-1}$ as its inputs. We call this new layer AttentionNeuron, and Figure~\ref{fig:arch} gives an overview of our method.
Inside AttentionNeuron, we model the observation $o_t$ as an arbitrarily ordered, variable-length list of sensory inputs, each of which is passed into its own \textit{sensory neuron}, a neural network module.
Each sensory neuron only has partial access to the agent's observation, at time $t$, the $i$th neuron can see only the $i$th component of the observation $o_t[i]$.
Combined with the previous action $a_{t-1}$, each sensory neuron computes messages $f_k(o_t[i], a_{t-1})$ and $f_v(o_t[i])$ that are broadcast to the rest of the system.
We then use attention to aggregate these messages into a \textit{global latent code}, $m_t$, that is PI with respect to the inputs.

The operations inside AttentionNeuron can be described by the following two equations. For clarity, Table~\ref{tab:notation} summarizes the notations as well as the corresponding setups we used for the experiments.

\begin{gather}
    K(o_t, a_{t-1}) = \begin{bmatrix}
    f_k(o_t[1], a_{t-1}) \\
    \cdots \\
    f_k(o_t[N], a_{t-1}) \\
    \end{bmatrix} \in \mathcal{R}^{N \times d_{f_k}},
    V(o_t) = \begin{bmatrix}
    f_v(o_t[1]) \\
    \cdots \\
    f_v(o_t[N]) \\
    \end{bmatrix} \in \mathcal{R}^{N \times d_{f_v}} \label{eq:funcs} \\
    m_t = \sigma \big{(} \frac{ [Q W_q ] [K(o_t, a_{t-1}) W_k]^{\top}}{\sqrt{d_q}} \big{)} [ V(o_t) W_v ]
\label{eq:attention}
\end{gather}

Equation~\ref{eq:funcs} shows how each of the $N$ sensory neuron independently generates its messages $f_k$ and $f_v$, which are functions shared across all sensory neurons. Equation~\ref{eq:attention} shows the attention mechanism aggregate these messages. Note that although we could have absorbed the projection matrices $W_q, W_k, W_v$ into $Q, K, V$, we keep them in the equation to show explicitly the formulation. Equation~\ref{eq:attention} is almost identical to the simple definition of self-attention mentioned earlier. Following~\cite{set2019}, we make our $Q$ matrix a bank of fixed embeddings, rather than depend on the observation $o_t$.


Note that permuting the observations only affects the row orders of $K$ and $V$, and that applying the same permutation to the rows of both $K$ and $V$ still results in the same $m_t$ which is PI. 
As long as we set constant the number of rows in $Q$, the change in the input size affects only the number of rows in $K$ and $V$ and does not affect the output $m_t$. In other words, our agent can accept inputs of arbitrary length and output a fixed sized $m_t$. Later, we apply this flexibility of input dimensions to RL agents.

\begin{table}[!htb]
\vskip -0.1in 
\centering
\captionof{table}{In this notation list, we provide the dimensions used in our model for different RL environments, to give the reader a sense of the relative magnitudes involved in each part of the system.}
\vskip 0.00in
\begin{small}
\begin{tabular}{llllll}
\hline
Description & Notation & CartPole                     & Ant                          & CarRacing                             & Atari Pong                            \\ \hline
Full observation space & $o_t$                  & $\mathcal{R}^5$              & $\mathcal{R}^{28}$             & $\mathcal{R}^{96 \times 96 \times 4}$ & $\mathcal{R}^{84 \times 84 \times 4}$ \\
Individual sensory input space & $o_t[i]$               & $\mathcal{R}^1$              & $\mathcal{R}^1$              & $\mathcal{R}^{6 \times 6 \times 4 = 144}$   & $\mathcal{R}^{6 \times 6 \times 4 = 144}$   \\
Number of sensory neurons & $N$                    & $5$                          & $28$                         & $(96 / 6)^2 = 256$                    & $(84 / 6)^2 = 196$                    \\
Dimension of action space & $|A|$                  & $1$                          & $8$                          & $3$                                   & $6$  (one-hot)               \\
Number of embeddings in $Q$ & $M$                    & $16$                         & $32$                         & $1024$                                & $400$                                 \\
Projection matrix for Q & $W_q$                  & $\mathcal{R}^{8 \times 32}$  & $\mathcal{R}^{8 \times 32}$  & $\mathcal{R}^{8 \times 16}$           & $\mathcal{R}^{8 \times 32}$           \\
Projection matrix for K & $W_k$                  & $\mathcal{R}^{8 \times 32}$  & $\mathcal{R}^{8 \times 32}$  & $\mathcal{R}^{111 \times 16}$         & $\mathcal{R}^{114 \times 32}$         \\
Projection matrix for V & $W_v$                  & $I$                          & $I$                          & $\mathcal{R}^{144 \times 16}$         & $\mathcal{R}^{144 \times 32}$         \\
Post-attention activation function& $\sigma(\cdot)$         & ${tanh}$     & ${tanh}$      & ${softmax}$   & ${softmax}$ \\
Global latent code & $m_t$                  & $\mathcal{R}^{16}$ & $\mathcal{R}^{32}$ & $\mathcal{R}^{1024 \times 16}$        & $\mathcal{R}^{400 \times 32}$         \\ \hline
\end{tabular}
\end{small}
\label{tab:notation}
\vskip -0.1in 
\end{table}

\subsection{Design Choices}
\vskip -0.1in 
It is worthwhile to have a discussion on the design choices made.
Since the ordering of the input is arbitrary, each sensory neuron is required to interpret and identify their received signal.
To achieve this, we want $f_k(o_t[i], a_{t-1})$ to have temporal memories.
In practice, we find both RNNs and feed-forward neural networks (FNN) with stacked observations work well, with FNNs being more practical for environments with high dimensional observations.

In addition to the temporal memory, including previous actions is important for the input identification too. Although the former allows the neurons to infer the input signals based on the characteristics of the temporal stream, this may not be sufficient. For example, when controlling a legged robot, most of the sensor readings are joint angles and velocities from the legs, which are not only numerically identically bounded but also change in similar patterns.
The inclusion of previous actions gives each sensory neuron a chance to infer the casual relationship between the input channel and the applied actions, which helps with the input identification.

Finally, in Equation~\ref{eq:attention} we could have combined $QW_q \in \mathcal{R}^{M \times d_q}$ as a single learnable parameters matrix, but we separate them for two reasons.
First, by factoring into two matrices, we can reduce the number of learnable parameters.
Second, we find that instead of making $Q$ learnable, using the positional encoding proposed in Transformer~\cite{vaswani2017} encourages the attention mechanism to generate distinct codes. Here we use the row indices in $Q$ as the positions for encoding.

\section{Experiments}
\vskip -0.1in 
We experiment on several different RL environments to study various properties of permutation invariant RL agents.
Due to the nature of the underlying tasks, we will describe the different architectures of the policy networks used and discuss various different training methods.
However, the AttentionNeuron layers in all agents are similar, so we first describe the common setups.
Hyper-parameters and other details for all experiments are summarized in Appendix~\ref{sec:setup_details}.

For non-vision continuous control tasks, the agent receives an observation vector $o_t \in \mathcal{R}^{|O|}$ at time $t$. We assign $N=|O|$ sensory neurons for the tasks, each of which sees one element from the vector, hence $o_t[i] \in \mathcal{R}^1, i=1, \cdots, |O|$. We use an LSTM~\cite{lstm1997} as our $f_k(o_t[i], a_{t-1})$ to generate Keys, the input size of which is $1 + |A|$ ($2$ for Cart-Pole and $9$ for PyBullet Ant). A simple pass-through function $f(x) = x$ serves as our $f_v(o_t[i])$, and $\sigma(\cdot)$ is $tanh$. For simplicity, we find $W_v = I$ works well for the tasks, so the learnable components are the LSTM, $W_q$ and $W_k$.

For vision based tasks, we gray-scale and stack $k=4$ consecutive RGB frames from the environment, and thus our agent observes $o_t \in \mathcal{R}^{H \times W \times k}$. 
$o_t$ is split into non-overlapping patches of size $P=6$ using a sliding window, so each sensory neuron observes $o_t[i] \in \mathcal{R}^{6 \times 6 \times k}$.
Here, $f_v(o_t[i])$ flattens the data and returns it, hence $V(o_t)$ returns a tensor of shape $N \times d_{f_v} = N \times (6 \times 6 \times 4) = N \times 144$. Due to the high dimensionality for vision tasks, we do not use RNNs for $f_k$, but instead use a simpler method to process each sensory input. $f_k(o_t[i], a_{t-1})$ takes the difference between consecutive frames ($o_t[i]$), then flattens the result, appends $a_{t-1}$, and returns the concatenated vector. $K(o_t, a_{t-1})$ thus gives a tensor of shape $N \times d_{f_k} = N \times [(6 \times 6 \times 3) + |A|] = N \times (108 + |A|)$ (111 for CarRacing and 114 for Atari Pong). We use ${softmax}$ as the non-linear activation function $\sigma(\cdot)$, and we apply layer normalization~\cite{ba2016layer} to both the input patches and the output latent code.

\subsection{Cart-pole swing up}
\vskip -0.1in 
We examine Cart-pole swing up~\cite{Gal2016Improving,deepPILCOgithub,ha2017evolving,wann2019} to first illustrate our method, and also use it to provide a clear analysis of the attention mechanism.
We use \texttt{CartPoleSwingUpHarder}~\cite{learningtopredict2019}, a more difficult version of the task where the initial positions and velocities are highly randomized, leading to a higher variance of task scenarios.
In the environment, the agent observes $[x, \dot{x}, cos(\theta), sin(\theta), \dot{\theta}]$, outputs a scalar action, and is rewarded at each step for getting $x$ close to 0 and $cos(\theta)$ close to 1.

\begin{figure}[!htb]
\vskip -0.10in 
\centering        
\includegraphics[width=1\textwidth]{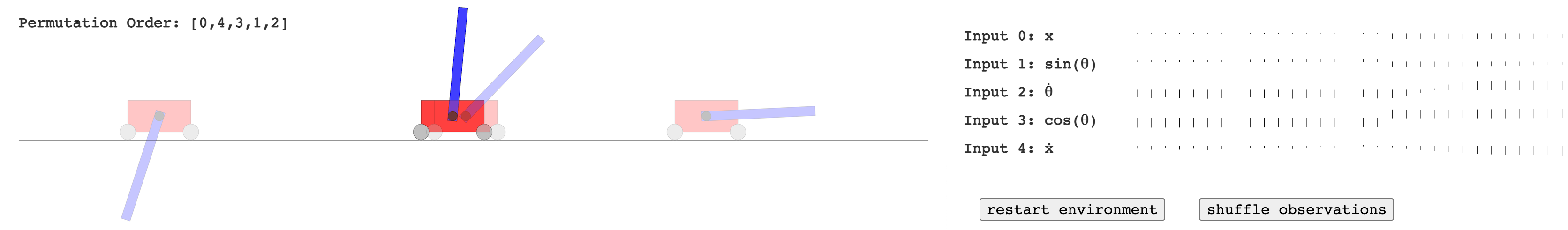}       
\vskip -0.05in 
\caption[In our web demo, the user can shuffle the order of the 5 inputs at any time, and observe how the agent adapts to the new ordering of the inputs.]{\textit{Interactive demo of CartPoleSwingUpHarder.} In our web demo\footnotemark, the user can shuffle the order of the 5 inputs at any time, and observe how the agent adapts to the new ordering of the inputs.}
\label{fig:cartpole_demo} 
\vskip -0.10in 
\end{figure}
\footnotetext{Interactive demos and videos of our results available at \url{https://attentionneuron.github.io/}}

We use a two-layer neural network as our agent. The first layer is an AttentionNeuron layer with $N=5$ sensory neurons and outputs $m_t \in \mathcal{R}^{16}$. A linear layer takes $m_t$ as input and outputs a scalar action. For comparison, we also trained an agent with a two-layer FNN policy with $16$ hidden units. We use direct policy search to train agents with CMA-ES~\cite{hansen2006cma}, an evolution strategies (ES) method.

\begin{table}[!htb]
\vskip -0.15in 
\centering
\caption{
\textit{Cart-pole Tests.} For each experiment, we report the average score and the standard deviation from 1000 test episodes. Our agent is trained only in the environment with 5 sensory inputs.
}
\label{tab:cartpole}
\vskip 0.08in 
\begin{tabular}{lllll}
\hline
                      & 5 obs                      & 5 obs (shuffled) &10 obs                           & 5 obs + 5 noise        \\ \hline
FNN (trained with 5 obs)   & $593 \pm 433$              & $38 \pm 120$ & N/A & N/A               \\
FNN (trained with 10 obs)  & N/A               & N/A & $593 \pm 433$       & $137 \pm 242$ \\
Ours (trained with 5 obs) & $472 \pm 426$ & $471 \pm 426$ & $471 \pm 425$       & $461 \pm 410$ \\ \hline
\end{tabular}
\vskip -0.15in 
\end{table}

Our agent can perform the task and balance the cart-pole from an initially random state.
Its average score is slightly lower than the baseline (See column 1 of Table~\ref{tab:cartpole}) because each sensory neuron requires some time steps in each episode to interpret the sensory input signal it receives. However, as a trade-off for the performance sacrifice, our agent can retain its performance even when the input sensor array is randomly shuffled, which is not the case for an FNN policy (column 2).
Moreover, although our agent is only trained in an environment with five inputs, it can accept an arbitrary number of inputs in any order without re-training\footnote{Because our agent was not trained with normalization layers, we scaled the output from the AttentionNeuron layer by 0.5 to account for the extra inputs in the last 2 experiments.}.
We test our agent by duplicating the 5 inputs to give the agent 10 observations (column 3).
When we replace the 5 extra signals with white noises with $\sigma=0.1$ (column 4), we do not see a significant drop in performance.

\begin{figure}[!htb]   
\vskip -0.10in 
\centering        
\includegraphics[width=1\textwidth]{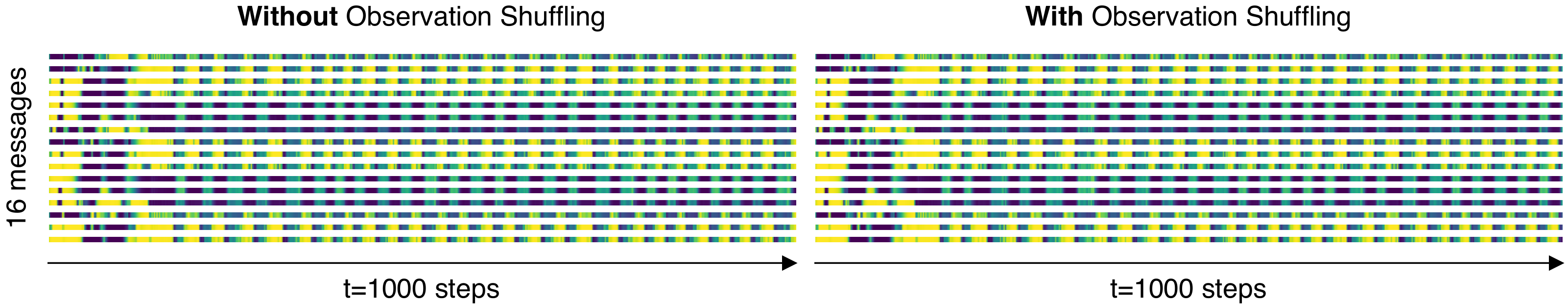}       
\vskip -0.1in 
\caption{
\textit{Permutation invariant outputs.} The output (16-dimensional global latent code) from the AttentionNeuron layer does not change when we input the sensor array as-is (left) or when we randomly shuffle the array (right). Yellow represents higher values, and blue for lower values.
}
\vskip -0.10in 
\label{fig:msg_analysis}
\end{figure}

The AttentionNeuron layer should possess 2 properties to attain these: its output is permutation invariant to its input, and its output carries task-relevant information.
Figure~\ref{fig:msg_analysis} is a visual confirmation of the permutation invariant property, whereby we plot the output messages from the layer and their changes over time from two tests. Using the same random seed, we keep the observation as-is in the first test but we shuffle the order in the second. As the figure shows, the output messages are identical in the two roll-outs.
We also perform a simple linear regression analysis on the outputs (based on the shuffled inputs) to recover the 5 inputs in their original order.
Table~\ref{tab:cartpole_analysis} shows the $R^2$ values\footnote{$R^2$ measures the goodness-of-fit of a model. An $R^2$ of 1 implies that the regression perfectly fits the data.} from this analysis, suggesting that some important indicators (e.g. $\dot{x}$ and $\dot{\theta}$) are well represented in the output.

\begin{minipage}[c]{0.5\textwidth}
\centering
\captionof{table}{\textit{Linear regression analysis on the output.}
\newline
For each of the $N=5$ sensory inputs we have one linear regression model with $m_t \in \mathcal{R}^{16}$ as the explanatory variables.}
\vskip -0.08in 
\begin{tabular}{llllll}
\hline
      & $x$     & $\dot{x}$ & $cos(\theta)$ & $sin(\theta)$ & $\dot{\theta}$ \\ \hline
$R^2$ & $0.354$ & $0.620$   & $0.626$       & $0.233$       & $0.550$        \\ \hline
\end{tabular}
\label{tab:cartpole_analysis}
\end{minipage}
\hspace{0.15in} 
\begin{minipage}[c]{0.335\textwidth}
\centering
\captionof{table}{\textit{PyBullet Ant results.}}
\vskip -0.08in 
\begin{tabular}{ll}
\hline
               & Score         \\ \hline
FNN (teacher)  & $2700 \pm 28$ \\
FNN (shuffled)  & $232 \pm 112$ \\
Ours (ES, shuffled)      & $2576 \pm 75$ \\ 
Ours (BC, shuffled) & $2034 \pm 948$ \\
Ours (BC, shuffled, larger) & $2579 \pm 457$ \\
\hline
\end{tabular}
\label{tab:bulletant_results}
\end{minipage}

\subsection{PyBullet Ant}\label{sec:bullet_ant}
\vskip -0.05in 
While direct policy search methods such as evolution strategies (ES) can train permutation invariant RL agents, oftentimes we already have access to pre-trained agents or recorded human data performing the task at hand.
Behavior cloning (BC) can allow us to convert an existing policy to a version that is permutation invariant with desirable properties associated with it.

In Table~\ref{tab:bulletant_results}, we train a standard two-layer FNN policy to perform \texttt{AntBulletEnv-v0}, a 3D locomotion task in PyBullet~\cite{coumans2020}, and use it as a teacher for BC. For comparison, we also train a two-layer agent with AttentionNeuron for its first layer. Both networks are trained with ES.
Similar to CartPole, we expect to see a small performance drop due to some time steps required for the agent to interpret an arbitrarily ordered observation space.
We then collect data from the FNN teacher policy to train permutation invariant agents using BC. More details of the BC setup can be found in Appendix~\ref{sec:discussion_pybulletant}.

The performance of the BC agent is lower than the one trained from scratch with ES, despite having the identical architecture.
This suggests that the inductive bias that comes with permutation invariance may not match the original teacher network, so the small model used here may not be expressive enough to clone any teacher policy, resulting in a larger variance in performance. A benefit of gradient-based BC, compared to RL, is that we can easily train larger networks to fit the behavioral data. We show that increasing the size of the subsequent layers for BC does enhance the performance.

As we will demonstrate next, BC is a useful technique for training permutation invariant agents in environments with high dimensional visual observations that may require larger networks.



\subsection{Atari Pong}
\vskip -0.1in 
Here, we are interested in solving screen-shuffled versions of vision-based RL environments, where each observation frame is divided up into a grid of patches, and like a puzzle, the agent must process the patches in a shuffled order to determine a course of action to take. A shuffled version of Atari Pong~\cite{openai_gym} (See Figure~\ref{fig:cover_diagram}, right pair) can be especially hard for humans to play when inductive biases from human priors~\cite{dubey2018investigating} that expect a certain type of spatial structure is missing from the observations.

But rather than throwing away the spatial structure entirely from our solution, we find that convolution neural network (CNN) policies work better than fully connected multi-layer perceptron (MLP) policies when trained with behavior cloning for Atari Pong. In this experiment, we reshape the output $m_t$ of the AttentionNeuron layer from $\mathcal{R}^{400 \times 32}$ to $\mathcal{R}^{20 \times 20 \times 32}$, a 2D grid of latent codes, and pass this 2D grid into a CNN policy. This way, the role of the AttentionNeuron layer is to take a list of unordered observation patches, and learn to construct a 2D grid representation of the inputs to be used by a downstream policy that expects some form of spatial structure in the codes. Our permutation invariant policy trained with BC can consistently reach a perfect score of 21, even with shuffled screens. The details of the CNN policy and BC training can be found in Appendix~\ref{sec:discussion_pong}.


\begin{figure}[!htb]   
\vskip -0.0in 
\centering        
\includegraphics[width=1\textwidth]{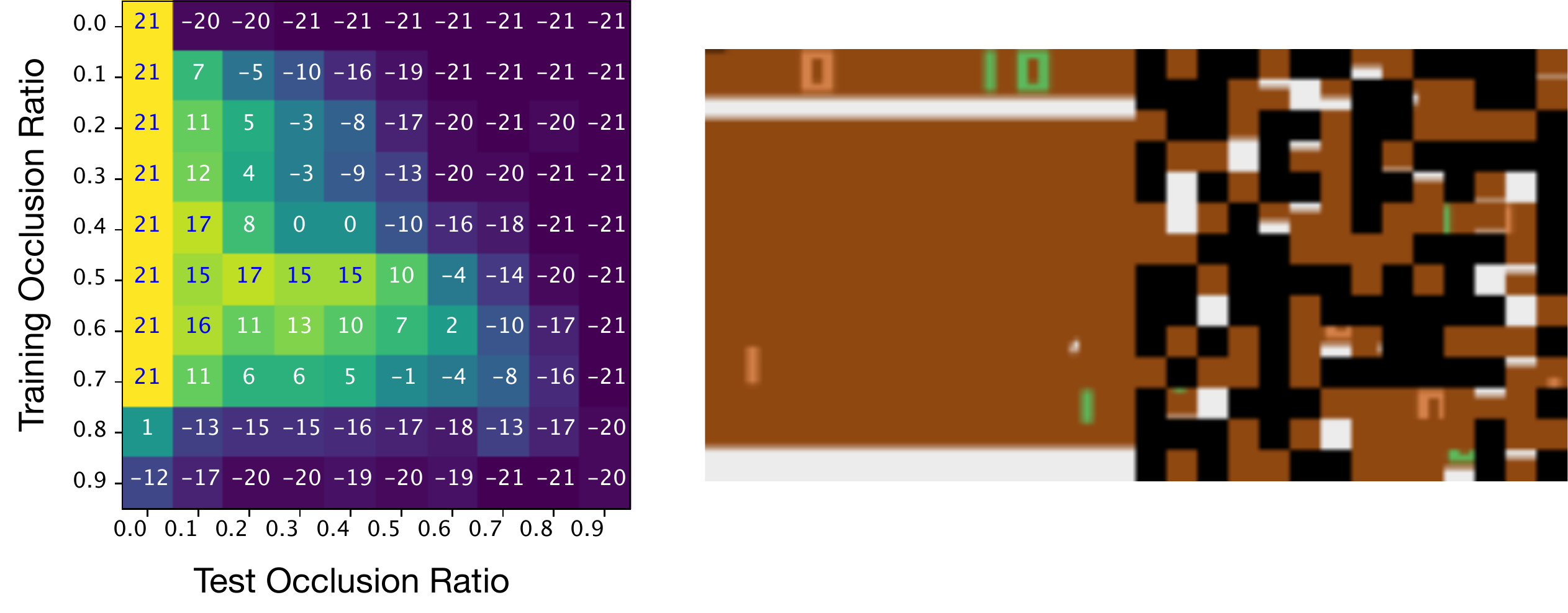} 
\vskip -0.05in 
\caption{
\textit{Mean test scores in Atari Pong, and example of a randomly-shuffled occluded observation.} In the heat map, each value is the average score from 100 test episodes. For comparison, we show the original screen (left), and the agent's observation (right). Discarded patches shown here in black.
}
\vskip -0.10in 
\label{fig:pong}
\end{figure}

Unlike typical CNN policies, our agent can accept a subset of the screen, since the agent's input is a variable-length list of patches.
It would thus be interesting to deliberately randomly discard a certain percentage of the patches and see how the agent reacts.
The net effect of this experiment for humans is similar to being asked to play a partially occluded and shuffled version of Atari Pong (see Figure~\ref{fig:pong}, right). During training via BC, we randomly remove a percentage of observation patches. In tests, we fix the randomly selected positions of patches to discard during an entire episode.

We present the results in a heat map in Figure~\ref{fig:pong} (left), where the y-axis shows the patches removed during training and the x-axis gives the patch occlusion ratio in tests.
The diagram shows clear patterns for interpretation.
Looking horizontally along each row, the performance drops because the agent sees less of the screen which increases the difficulty.
Interestingly, an agent trained at a high occlusion rate of $80\%$ rarely wins against the Atari opponent, but once it is presented with the full set of patches during tests, it is able to achieve a fair result by making use of the additional information.

\begin{figure}[ht]   
\vskip -0.05in 
\centering        
\includegraphics[width=1\textwidth]{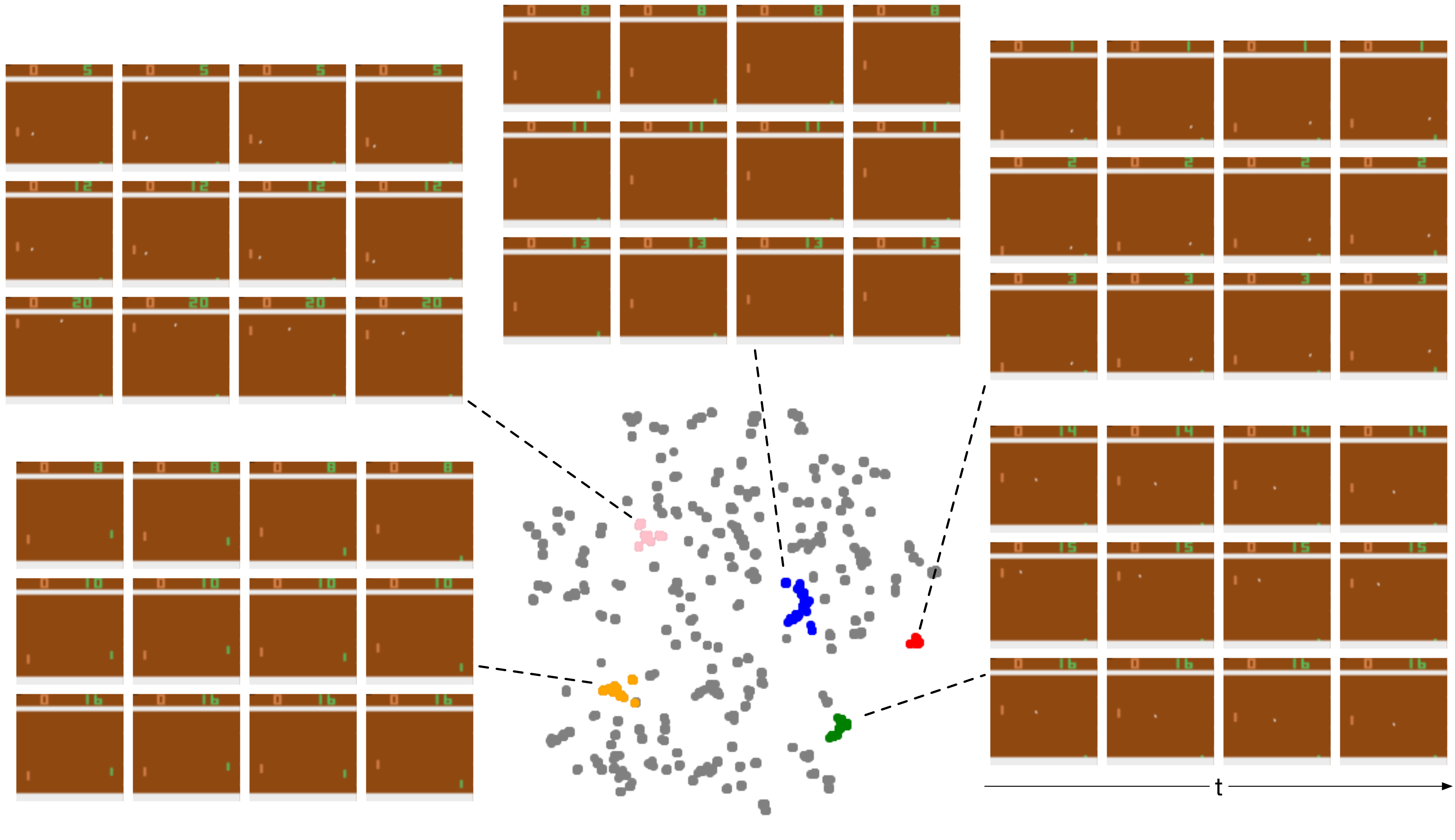}
\vskip -0.05in 
\caption{
\textit{2D embedding of the AttentionNeuron layer's output in a test episode.}
We highlight several representative groups in the plot, and show the sampled inputs from them.
For each group, we show 3 corresponding inputs (rows) and unstack each to show the time dimension (columns). 
}
\vskip -0.2in 
\label{fig:pong_analysis}
\end{figure}

To gain insights into understanding the policy, we projected the AttentionNeuron layer's output in a test roll-out to 2D space using t-SNE~\cite{van2008visualizing}. In Figure~\ref{fig:pong_analysis}, we highlight several groups and show their corresponding inputs. The AttentionNeuron layer clearly learned to cluster inputs that share similar features. For example, the 3 sampled inputs in the blue group show the situation when the agent's paddle moved toward the bottom of the screen and stayed there. Similarly, the orange group shows the cases when the ball was not in sight, this happened right before/after a game started/ended. We believe these discriminative outputs enabled the downstream policy to accomplish the agent's task.

\subsection{CarRacing}
\vskip -0.1in 

We find that encouraging an agent to learn a coherent representation of a deliberately shuffled visual scene leads to agents with useful generalization properties.
Such agents are still able to perform their task even if the visual background of the environment changes, despite being trained only on a single static background.
Out-of-domain generalization is an active area, and here, we combine our method with AttentionAgent~\cite{attentionagent2020}, a method that uses selective, hard-attention via a patch voting mechanism. AttentionAgents in \cite{attentionagent2020} generalize well to several unseen visual environments where task irrelevant elements are modified, but fail to generalize to drastic background changes in a zero-shot setting.

In this experiment, we combine the permutation invariant AttentionNeuron layer with the policy network used in AttentionAgent. As their hard-attention-based policy is non-differentiable, we train the entire system using ES.
We reshape the AttentionNeuron layer's outputs to adapt for the policy network.
Specifically, we reshape the output message to $m_t \in \mathcal{R}^{32 \times 32 \times 16}$ such that it can be viewed as a 32-by-32 grid of 16 channels.
The end result is a policy with two layers of attention: the first layer outputs a latent code book to represent a shuffled scene, and the second layer performs hard attention to select the top $K=10$ codes from a 2D global latent code book. A detailed description of the selective hard attention policy from \cite{attentionagent2020} and other training details can be found in Appendix~\ref{sec:discussion_carracing}.

We first train the agent in the CarRacing~\cite{carracing_v0} environment, and report the average score from 100 test roll-outs in Table~\ref{tab:carracing_results}. As the first column shows, our agent's performance in the training environment is slightly lower but comparable to the baseline method, as expected. But because our agent accepts randomly shuffled inputs, it is still able to navigate even when the patches are shuffled. Figure~\ref{fig:cover_diagram} (left pair) gives an illustration, where the right screen is what our agent observes and the left is for human visualization. A human will find driving with the shuffled observation to be very difficult because we are not constantly exposed to such tasks, just like in the ``reverse bicycle'' example mentioned earlier.

\begin{table}[!htb]
\vskip -0.15in 
\centering
\begin{small}
\captionof{table}{CarRacing Test Results.}
\vskip 0.05in 
\begin{tabular}{llllll}
\hline
               & Training Env  & KOF           & Mt. Fuji     & Ukiyoe        & DS            \\ \hline
AttentionAgent~\cite{attentionagent2020} & $901 \pm 54$  & $-81 \pm 4$   & $-57 \pm 38$ & $-107 \pm 50$ & $-56 \pm 23$  \\
NetRand~\cite{lee2019network} & $480 \pm 144$ & $20 \pm 84$ & $356 \pm 159$ & $533 \pm 111$ & $-27 \pm 34$ \\
NetRand + AttentionAgent & $885 \pm 64$ & $-51 \pm 14$ & $709 \pm 94$ & $656 \pm 131$ & $122 \pm 134$ \\
Ours        & $801 \pm 147$ & $646 \pm 189$ & $503 \pm 152$ & $661 \pm 140$ & $171 \pm 146$ \\ \hline

\end{tabular}
\label{tab:carracing_results}
\end{small}
\vskip -0.05in 
\end{table}

\begin{figure}[ht]   
\vskip -0.05in 
\centering        
\includegraphics[width=1\textwidth]{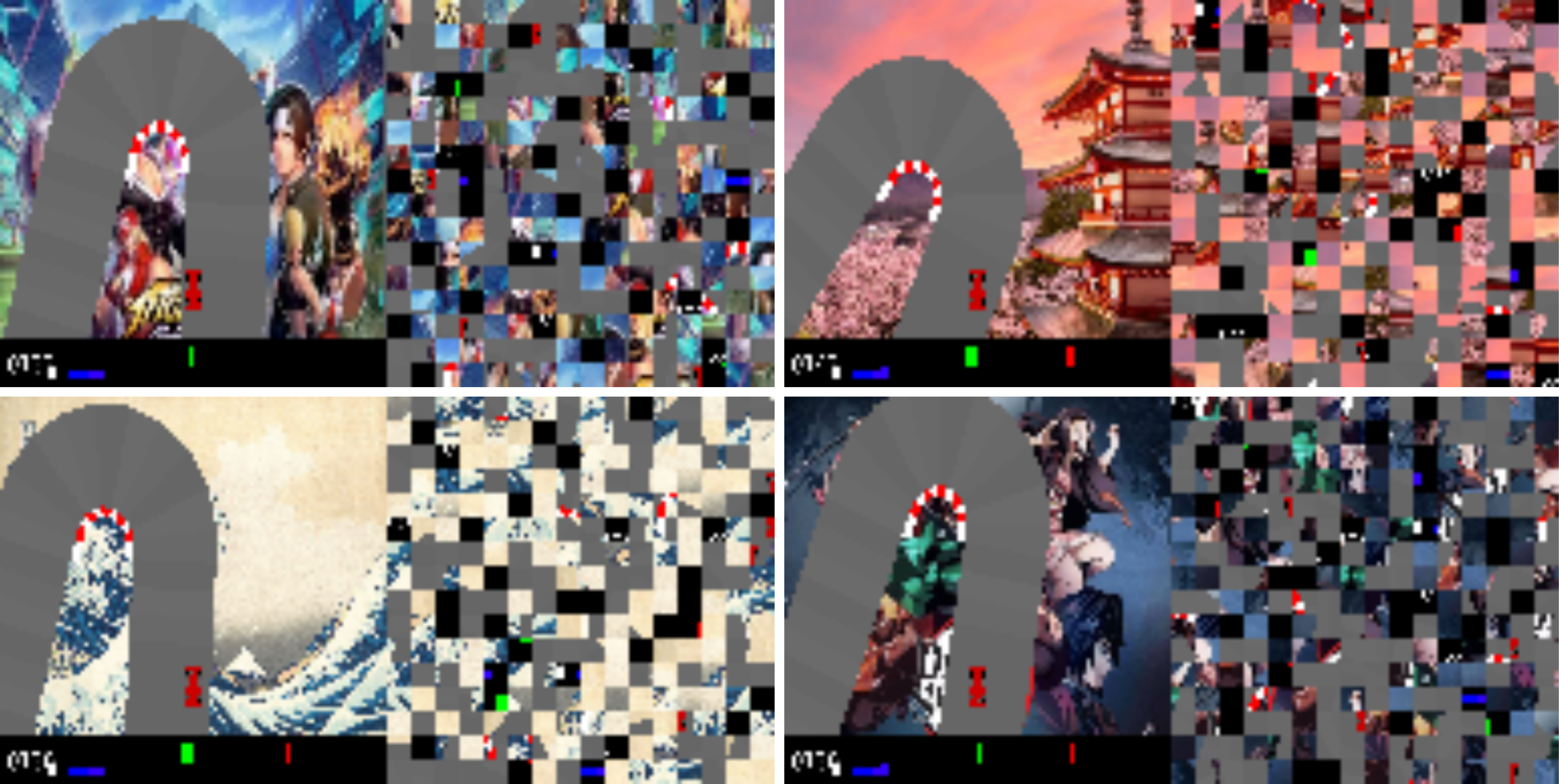}
\vskip -0.05in 
\caption{
\textit{Screenshots of test environments.} In each pair of images, the left is for human visualization and the right is what our agent sees. From the top left and in the clockwise order, the environments are ``KOF'', ``Mt. Fuji'', ``DS'' and ``Ukiyoe''.
}
\vskip -0.25in 
\label{fig:carracing_bkg}
\end{figure}

Without additional training or fine-tuning, we test whether the agent can also navigate in four modified environments where the green grass background is replaced with various images (See Figure~\ref{fig:carracing_bkg}). As Table~\ref{tab:carracing_results} (from column 2) shows, our agent generalizes well to most of the test environments with only mild performance drops while the baseline method fails to generalize. We suspect this is because the AttentionNeuron layer has transformed the original RGB space to a useful hidden representation (represented by $m_t$) that has eliminated task irrelevant information after observing and reasoning about the sequences of $(o_t, a_{t-1})$ during training, enabling the downstream hard attention policy to work with an optimized abstract representation tailored for the policy, instead of raw RGB patches.

We also compare our method to NetRand~\cite{lee2019network}, a simple but effective technique developed to perform similar generalization tasks. In the second row of Table~\ref{tab:carracing_results} are the results of training NetRand on the base CarRacing task. The CarRacing task proved to be too difficult for NetRand, but despite a low performance score of 480 in the training environment, the agent generalizes well to the ``Mt. Fuji'' and ``Ukiyoe'' modifications. In order to obtain a meaningful comparison, we combine NetRand with AttentionAgent so that it can get close to a mean score of 900 on the base task. To do that, we use NetRand as an input layer to the AttentionAgent policy network, and train the combination end-to-end using ES, which is consistent with our proposed method for this task. The combination attains a respectable mean score of 885, and as we can see in the third row of the above table, this approach also generalizes to a few of the unseen modifications of the CarRacing environment.

Our score on the base CarRacing task is lower than NetRand, but this is expected since our agent requires some amount of time steps to identify each of the inputs (which could be shuffled), while the NetRand and AttentionAgent agent will simply fail on the shuffled versions of CarRacing. Despite this, our method still compares favorably on the generalization performance.

\begin{figure}[ht]   
\vskip -0.15in 
\centering        
\includegraphics[width=1\textwidth]{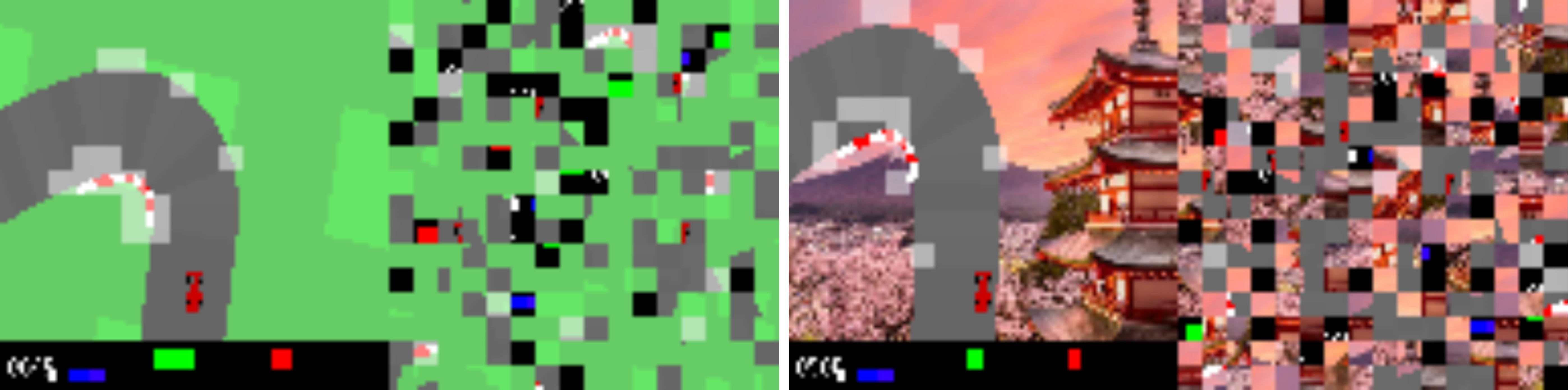}
\vskip -0.05in 
\caption{
\textit{Attention visualization.} We highlight the observed patches that receive the most attention. Left: Training environment. Right: Test environment with unseen background.
}
\vskip -0.10in 
\label{fig:attention_viz}
\end{figure}

We visualize the attentions from the AttentionNeuron layer in Figure~\ref{fig:attention_viz}.
In CarRacing, the agent has learned to focus its attention (indicated by the highlighted patches) on the road boundaries which are intuitive to human beings and are critical to the task. Notice that the attended positions are consistent before and after the shuffling. More details about this visualization can be found in Appendix~\ref{sec:discussion_carracing}.

\section{Discussion and Future Work}\label{sec:discussion}
\vskip -0.1in 

In this work, we investigate the properties of RL agents that can treat their observations as an arbitrarily ordered, variable-length list of sensory inputs. By processing each input stream independently, and consolidating the processed information using attention, our agents can still perform their tasks even if the ordering of the observations is randomly permuted several times during an episode, without explicitly training for frequent re-shuffling (See Table~\ref{tab:reshuffling}).

\begin{table}[!htb]
\vskip -0.1in
\centering
\caption{\textit{Reshuffle observations during a roll-out.}
In each test episode, we reshuffle the observations every $t$ steps.
For CartPole, we test for 1000 episodes because of its larger task variance. For the other tasks, we report mean and standard deviation from 100 tests.  All environments except for Atari Pong have a hard limit of 1000 time steps per episode. In Atari Pong, while the maximum length of an episode does not exist, we observed that an episode usually lasts for around 2500 time steps.
}
\vskip 0.05in
\begin{tabular}{lllll}
\hline
             & CartPole      & PyBullet Ant   & Atari Pong   & CarRacing     \\ \hline
$t=25$       & $107 \pm 146$ & $2053 \pm 225$ & $-20 \pm 1$  & $732 \pm 161$ \\ 
$t=50$       & $163 \pm 198$ & $2319 \pm 188$ & $-20 \pm 2$  & $772 \pm 163$ \\ 
$t=100$      & $242 \pm 254$ & $2406 \pm 178$ & $-10 \pm 12$ & $768 \pm 167$ \\ 
$t=200$      & $318 \pm 310$ & $2493 \pm 105$ & $-2 \pm 17$  & $774 \pm 182$ \\ 
$t=500$      & $407 \pm 380$ & $2548 \pm 87$  & $18 \pm 9$   & $805 \pm 158$ \\ 
No reshuffle & $472 \pm 426$ & $2576 \pm 75$  & $21 \pm 0$   & $801 \pm 147$ \\ \hline 
\end{tabular}
\label{tab:reshuffling}
\vskip -0.25in
\end{table}

\textbf{Applications}\: By presenting the agent with shuffled, and even incomplete observations, we encourage it to interpret the meaning of each local sensory input and how they relate to the global context.
This could be useful in many real world applications. For example, such policies could avoid errors due to cross-wiring or complex, dynamic input-output mappings when being deployed in real robots. A similar setup to the CartPole experiment with extra noisy channels could enable a system that receives thousands of noisy input channels to identify the small subset of channels with relevant information.

\textbf{Limitations}\: For visual environments, patch size selection will affect both performance and computing complexity. We find that patches of 6x6 pixels work well for our tasks, as did 4x4 pixels to some extent, but single pixel observations fail to work. Small patch sizes also result in a large attention matrix which may be too costly to compute, unless approximations are used~\cite{wang2020linformer,choromanski2020rethinking,xiong2021nystr}.

Another limitation is that the permutation invariant property applies only to the inputs, and not to the outputs. While the ordering of the observations can be shuffled, the ordering of the actions cannot. For permutation invariant outputs to work, each action will require feedback from the environment, including reward information, in order to learn the relationship between itself and the environment.

\textbf{Societal Impact}\: Like most algorithms proposed in computer science and machine learning, our method can be applied in ways that will have potentially positive or negative impacts to society. While our small-scale, self-contained experiments study only the properties of agents that are PI to their observations, and we believe our results do not directly cause harm to society, the robustness and flexible properties of the method may be of use for data-collection systems that receive data from a large variable number of sensors. For instance, one could apply permutation invariant sensory systems to process data from millions of sensors for anomaly detection, which may lead to both positive or negative impacts, if used in applications such as large-scale sensor analysis for weather forecasting, or deployed in large-scale surveillance systems that could undermine our basic freedoms.

Our work also provides a way to view the Transformer~\cite{vaswani2017} through the lens of self-organizing neural networks. Transformers are known to have potentially negative societal impacts highlighted in studies about possible data-leakage and privacy vulnerabilities~\cite{carlini2020extracting}, malicious misuse and issues concerning bias and fairness~\cite{bender2021dangers}, and energy requirements for training them~\cite{strubell2019energy}.

\textbf{Future Work}\: An interesting future direction is to also make the action layer have the same properties, and model each ``motor neuron'' as a module connected using attention. With such methods, it may be possible to train an agent with an arbitrary number of legs, or control robots with different morphology using a single policy that is also provided with a reward signal as feedback.
We look forward to seeing future works that include signals such as environmental rewards to train PI meta-learning agents that can adapt to not only changes in the observed environment, but also to changes to itself.


\section*{Acknowledgements}\label{sec:acknowledgements}
The authors would like to thank Rishabh Agarwal, Jie Tan, Yingtao Tian, Douglas Eck, Aleksandra Faust and our NeurIPS2021 reviewers for valuable discussion and feedback. The experiments in this work were conducted using virtual machines provided by Google Cloud Platform.


\appendix

\section{Appendix}
\subsection{Intuitive explanation of Self-Attention's permutation invariant property}
\label{sec:pi_explanation}
Here, we provide a simple, non-rigorous example demonstrating permutation invariant property of the self-attention mechanism, to give some intuition to readers who may not be familiar with self-attention. For a detailed treatment, please refer to \cite{set2019}.

As mentioned in Section~\ref{sec:method_background}, in its simplest form, self-attention is described as:
\begin{equation}
y = \sigma(QK^{\top})V
\end{equation}
where $Q \in \mathcal{R}^{N_q \times d_q}, K \in \mathcal{R}^{N \times d_q}, V \in \mathcal{R}^{N \times d_v}$ are the Query, Key and Value matrices and $\sigma(\cdot)$ is a non-linear function. In this work, $Q$ is a fixed matrix, and $K, V$ are functions of the input $X \in \mathcal{R}^{N \times d_{in}}$ where $N$ is the number of observation components (equivalent to the number of sensory neurons) and $d_{in}$ is the dimension of each component. In most settings, $K=X W_k, V=X W_v$ are linear transformations, thus permuting $X$ therefore is equivalent to permuting the rows in $K, V$.

We would like to show that the output $y$ is the same regardless of the ordering of the rows of $K, V$. For simplicity, suppose $N=3, N_q=2, d_q=d_v=1$, so that $Q \in \mathcal{R}^{2 \times 1}, K \in \mathcal{R}^{3 \times 1}, V \in \mathcal{R}^{3 \times 1}$:
\begin{equation}
\label{eq:before_permutation}
    \begin{aligned}y &= \sigma \Bigg( \begin{pmatrix} q_{1} \\ q_{2} \end{pmatrix} \left( k_{1}\; k_{2}\; k_{3}\right) \Bigg) \begin{pmatrix} v_{1} \\ v_{2} \\ v_{3} \end{pmatrix}\\
    &=  \sigma \Bigg( \begin{pmatrix} q_1 k_1 \;\; q_1 k_2 \;\; q_1 k_3 \\ q_2 k_1 \;\; q_2 k_2 \;\; q_2 k_3 \end{pmatrix} \Bigg)\begin{pmatrix} v_{1} \\ v_{2} \\ v_{3} \end{pmatrix}\\
    &= \begin{pmatrix} \textcolor{red}{\sigma (q_1 k_1) v_1} \;+\; \textcolor{blue}{\sigma (q_1 k_2) v_2} \;+\; \textcolor{orange}{\sigma (q_1 k_3) v_3} \\ \textcolor{green}{\sigma (q_2 k_1) v_1} \;+\; \textcolor{gray}{\sigma (q_2 k_2) v_2} \;+\; \textcolor{brown}{\sigma (q_2 k_3) v_3} \end{pmatrix}
    \end{aligned}
\end{equation}
The output $y \in \mathcal{R}^{2 \times 1}$ remains the same when the rows of $K, V$ are permuted from $[1, 2, 3]$ to $[3, 1, 2]$:
\begin{equation}
\label{eq:after_permutation}
    \begin{aligned}y &= \sigma \Bigg( \begin{pmatrix} q_{1} \\ q_{2} \end{pmatrix} \left( k_{3}\; k_{1}\; k_{2}\right) \Bigg) \begin{pmatrix} v_{3} \\ v_{1} \\ v_{2} \end{pmatrix}\\
    &=  \sigma \Bigg( \begin{pmatrix} q_1 k_3 \;\; q_1 k_1 \;\; q_1 k_2 \\ q_2 k_3 \;\; q_2 k_1 \;\; q_2 k_2 \end{pmatrix} \Bigg)\begin{pmatrix} v_{3} \\ v_{1} \\ v_{2} \end{pmatrix}\\
    &= \begin{pmatrix} \textcolor{orange}{\sigma (q_1 k_3) v_3} \;+\; \textcolor{red}{\sigma (q_1 k_1) v_1} \;+\; \textcolor{blue}{\sigma (q_1 k_2) v_2} \\ \textcolor{brown}{\sigma (q_2 k_3) v_3} \;+\; \textcolor{green}{\sigma (q_2 k_1) v_1} \;+\; \textcolor{gray}{\sigma (q_2 k_2) v_2} \end{pmatrix}
    \end{aligned}
\end{equation}
We have highlighted the same terms with the same color in Equations~\ref{eq:before_permutation} and~\ref{eq:after_permutation} to show the results are indeed identical. In general, we have $y_{ij} = \sum_{b=1}^{N} \sigma \big{(} \sum_{a=1}^{d_q} Q_{ia} K_{ba} \big{)} V_{bj}$. Permuting the input is equivalent to permuting the indices $b$ (i.e. rows of $K$ and $V$), which only affects the order of the outer summation and does not affect $y_{ij}$ because summation is a permutation invariant operation. Notice that in the above example and the proof here we have assumed that $\sigma(\cdot)$ is an element-wise operation---a valid assumption since most activation functions satisfy this condition.\footnote{Applying \textit{softmax} to each row only brings scalar multipliers to each row and the proof still holds.}

As discussed in Section~\ref{sec:method_actual}, this formulation lets us convert an observation signal from the RL environment into a permutation invariant representation $y$. We can use this representation in place of the actual observation as the input that goes into the downstream policy network of an RL agent.


\newpage
\subsection{Hyper-parameters}
\label{sec:hp}

Table~\ref{tab:notation} contains the hyper-parameters used for each experiment. We did not employ exhaustive hyper-parameter tuning, but have simply selected parameters that can appropriately size our models to work with training methods such as evolution strategies, where the number of parameters cannot be too large. As mentioned in the discussion section about the limitations, we tested a small range of patch sizes (1 pixel, 4 pixels, 6 pixels), and we find that a patch size of 6x6 works well across tasks.

\subsection{Description of compute infrastructure used to conduct experiments}
\label{sec:infra}

For all ES results, we train on Google Kubernetes Engines (GKE) with 256 CPUs (N1 series) for each job.
The approximate time, including both training and periodic tests, for the jobs are: 3 days (CartPole), 5 days (PyBullet Ant ES) and 10 days (CarRacing).
For BC results, we train with Google Computing Engines (GCE) on an instance that has one V100 GPU.
The approximate time, including both training and periodic tests, for the jobs are: 5 days (PyBullet Ant BC), 1 day (Atari Pong).

\subsection{Detailed setups for the experiments}
\label{sec:setup_details}

\subsubsection{Training budget}
\label{sec:discussion_training}

The costs of ES training are summarized in the following table. A maximum of 20K generations is specified in the training, but stopped early if the performance converged. Each generation has $256 \times 16=4096$ episode rollouts, where $256$ is the population size and $16$ is the rollout repetitions. The Pong permutation-invariant (PI) agents were trained using behavior cloning (BC) on a pre-trained PPO policy (which is not PI-capable), with 10M training steps.

\begin{table}[h!]
\begin{tabular}{lllll}
\hline
Environment              & CartPoleSwingUpHarder & PyBullet Ant & Atari Pong & CarRacing \\ \hline
Number of Generations & 14,000      & 12,000          & -          & 4,000        \\ \hline
\end{tabular}
\end{table}

Note that we used the hyper-parameters (e.g., population size, rollout repetitions) that proved to work on a wide range of tasks from past experience, and did not tune them for each experiment. In other words, these settings were not chosen with sample-efficiency in mind, but rather for learning a working PI-capable policy using distributed computation within a reasonable wall-clock time budget.

We consider two possible approaches when we take sample-efficiency into consideration.
In the experiments, we have demonstrated that it is possible to simply use state-of-the-art RL algorithms to learn a non-PI policy, and then use BC to produce a PI version of the policy.
The first approach is thus to rely on the conventional RL algorithms to increase sample efficiency, which is a hot and on-going topic in the area. On the other hand, we do think that an interesting future direction is to formulate environments where BC will fail in a PI setting, and that interactions with the environment (in a PI setting) is required to learn a PI policy. For instance, we have demonstrated in PyBullet Ant that the BC method requires the cloned agent to have a much larger number of parameters compared to one trained with RL. This is where an investigation in sample-efficiency improvements in the RL algorithm explicitly in the PI setting may be beneficial.

\subsubsection{PyBullet Ant}
\label{sec:discussion_pybulletant}

In the PyBullet Ant experiment, we demonstrated that a pre-trained policy can be converted into a permutation invariant one with behavior cloning (BC). We give detailed task description and experimental setups here. In \texttt{AntBulletEnv-v0}, the agent controls an ant robot that has 8 joints ($|A| = 8$), and gets to see an observation vector that has base and joint states as well as foot-ground contact information at each time step (|O|=28). The mission is to make the ant move along a pre-defined straight line as fast as possible.
The teacher policy is a 2-layer FNN policy that has 32 hidden units trained with ES. We collected data from 1000 test roll-outs, each of which lasted for 500 steps. During training, we add zero-mean Gaussian noise ($\sigma=0.03$) to the previous actions. For the student policy, We set up two networks. The first policy is a 2-layered network that has the AttentionNeuron with output size $m_t \in \mathcal{R}^{32}$ as its first layer, followed by a fully-connected (FC) layer. The second, larger policy is similar in architecture, but we added one more FC layer and expanded all hidden size to $128$ to increase its expressiveness. We train the students with a batch size of $64$, an Adam optimizer of $lr=0.001$ and we clip the gradient at maximum norm of $0.5$.

\subsubsection{Atari Pong}
\label{sec:discussion_pong}

In the Atari game Pong, we append a deep CNN to the AttentionNeuron layer in our agent (student policy). To be concrete, we reshape the AttentionNeuron's output message $m_t \in \mathcal{R}^{400 \times 32}$ to $m_t \in \mathcal{R}^{20 \times 20 \times 32}$ and pass it to the trailing CNN: [Conv(in=32, out=64, kernel=4, stride=2), Conv(in=64, out=64, kernel=3, stride=1), FC(in=3136, out=512), FC(in=512, out=6)]. We use $ReLU$ as the activation functions in the CNN. We collect the stacked observations and the corresponding logits output from a pre-trained PPO agent (teacher policy) from 1000 roll-outs, and we minimize the MSE loss between the student policy's output and the teacher policy's logits. The learning rate and norm clip are the same as the previous experiment, but we use a batch size of $256$.

For the occluded Pong experiment, we randomly remove a certain percentage of the patches across a training batch of stacked observation patches. In tests, we sample a patch mask to determine the positions to occlude at the beginning of the episode, and apply this mask throughout the episode.

\subsubsection{CarRacing}
\label{sec:discussion_carracing}

In AttentionAgent~\cite{attentionagent2020}, the authors observed that the agent generalizes well if it is forced to make decisions based on only a fraction of the available observations. Concretely, \cite{attentionagent2020} proposed to segment the input image into patches and let the patches vote for each other via a modified self-attention mechanism. The agent would then take into consideration only the top $K=10$ patches that have the most votes and based on the coordinates of which an LSTM controller makes decisions. Because the voting process involves sorting and pruning that are not differentiable, the agent is trained with ES. In their experiments, the authors demonstrated that the agent could navigate well not only in the training environment, but also zero-shot transfer to several modified environments.

We need only to reshape the AttentionNeuron layer's outputs to adapt for AttentionAgent's policy network. Specifically, we reshape the output message $m_t \in \mathcal{R}^{1024 \times 16}$ to $m_t \in \mathcal{R}^{32 \times 32 \times 16}$ such that it can be viewed as a 32-by-32 ``image'' of 16 channels. Then if we make AttentionAgent's patch segmentation size 1, the original patch voting becomes voting among the $m_t$'s and thus the output fits perfectly into the policy network. Except for this patch size, we kept all hyper-parameters in AttentionAgent unchanged, we also used the same CMA-ES training hyper-parameters.

Although the simple settings above allows our augmented agent to learn to drive and generalize to unseen background changes, we found the car jittered left and right through the courses. We suspect this is because of the frame differential operation in our $f_k(o_t, a_{t-1})$. Specifically, even when the car is on a straight lane, constantly steering left and right allows $f_k(o_t, a_{t-1})$ to capture more meaningful signals related to the changes of the road. To avoid such jittering behavior, we make $m_t$ a rolling average of itself: $m_t = (1 - \alpha)m_t + \alpha m_{t-1}, 0 \le \alpha \le 1$. In our implementation $\alpha = g([h_{t-1}, a_{t-1}])$, where $h_{t-1}$ is the hidden state from AttentionAgent's LSTM controller and $a_{t-1}$ is the previous action. $g(\cdot)$ is a 2-layer FNN with 16 hidden units and a $sigmoid$ output layer.

We analyzed the attention matrix in the AttentionNeuron layer and visualized the attended positions.
To be concrete, in CarRacing, the Query matrix has $1024$ rows.
Because we have $16 \times 16=256$ patches, the Key matrix has $256$ rows, we therefore have an attention matrix of size $1024 \times 256$.
To plot attended patches, we select from each row in the attention matrix the patch that has the largest value after softmax, this gives us a vector of length $1024$.
This vector represents the patches each of the $1024$ output channels has considered to be the most important.
$1024$ is larger than the total patch count, however there are duplications (i.e. multiple output channels have mostly focused on the same patches). The unique number turns out to be $10 \sim 20$ at each time step. We emphasize these patches on the observation images to create an animation.

\newpage

\bibliography{main}

\begin{thebibliography}{10}

\bibitem{amos2018differentiable}
B.~Amos, I.~D.~J. Rodriguez, J.~Sacks, B.~Boots, and J.~Z. Kolter.
\newblock Differentiable mpc for end-to-end planning and control.
\newblock {\em arXiv preprint arXiv:1810.13400}, 2018.

\bibitem{ba2016using}
J.~Ba, G.~Hinton, V.~Mnih, J.~Z. Leibo, and C.~Ionescu.
\newblock Using fast weights to attend to the recent past.
\newblock {\em arXiv preprint arXiv:1610.06258}, 2016.

\bibitem{ba2016layer}
J.~L. Ba, J.~R. Kiros, and G.~E. Hinton.
\newblock Layer normalization.
\newblock {\em arXiv preprint arXiv:1607.06450}, 2016.

\bibitem{bach1969vision}
P.~Bach-y Rita, C.~C. Collins, F.~A. Saunders, B.~White, and L.~Scadden.
\newblock Vision substitution by tactile image projection.
\newblock {\em Nature}, 221(5184):963--964, 1969.

\bibitem{bach2003sensory}
P.~Bach-y Rita and S.~W. Kercel.
\newblock Sensory substitution and the human--machine interface.
\newblock {\em Trends in cognitive sciences}, 7(12):541--546, 2003.

\bibitem{bender2021dangers}
E.~M. Bender, T.~Gebru, A.~McMillan-Major, and S.~Shmitchell.
\newblock On the dangers of stochastic parrots: Can language models be too big?
\newblock In {\em Proceedings of the 2021 ACM Conference on Fairness,
  Accountability, and Transparency}, pages 610--623, 2021.

\bibitem{openai_gym}
G.~Brockman, V.~Cheung, L.~Pettersson, J.~Schneider, J.~Schulman, J.~Tang, and
  W.~Zaremba.
\newblock Openai gym.
\newblock {\em arXiv preprint arXiv:1606.01540}, 2016.

\bibitem{brown2020language}
T.~B. Brown, B.~Mann, N.~Ryder, M.~Subbiah, J.~Kaplan, P.~Dhariwal,
  A.~Neelakantan, P.~Shyam, G.~Sastry, A.~Askell, et~al.
\newblock Language models are few-shot learners.
\newblock {\em arXiv preprint arXiv:2005.14165}, 2020.

\bibitem{carlini2020extracting}
N.~Carlini, F.~Tramer, E.~Wallace, M.~Jagielski, A.~Herbert-Voss, K.~Lee,
  A.~Roberts, T.~Brown, D.~Song, U.~Erlingsson, et~al.
\newblock Extracting training data from large language models.
\newblock {\em arXiv preprint arXiv:2012.07805}, 2020.

\bibitem{chang2020decentralized}
M.~Chang, S.~Kaushik, S.~M. Weinberg, T.~Griffiths, and S.~Levine.
\newblock Decentralized reinforcement learning: Global decision-making via
  local economic transactions.
\newblock In {\em International Conference on Machine Learning}, pages
  1437--1447. PMLR, 2020.

\bibitem{cheney2014unshackling}
N.~Cheney, R.~MacCurdy, J.~Clune, and H.~Lipson.
\newblock Unshackling evolution: evolving soft robots with multiple materials
  and a powerful generative encoding.
\newblock {\em ACM SIGEVOlution}, 7(1):11--23, 2014.

\bibitem{choi2017multi}
J.~Choi, B.-J. Lee, and B.-T. Zhang.
\newblock Multi-focus attention network for efficient deep reinforcement
  learning.
\newblock {\em arXiv preprint arXiv:1712.04603}, 2017.

\bibitem{chopard1998cellular}
B.~Chopard and M.~Droz.
\newblock {\em Cellular automata}, volume~1.
\newblock Springer, 1998.

\bibitem{choromanski2020rethinking}
K.~Choromanski, V.~Likhosherstov, D.~Dohan, X.~Song, A.~Gane, T.~Sarlos,
  P.~Hawkins, J.~Davis, A.~Mohiuddin, L.~Kaiser, et~al.
\newblock Rethinking attention with performers.
\newblock {\em arXiv preprint arXiv:2009.14794}, 2020.

\bibitem{chua1988cellular}
L.~O. Chua and L.~Yang.
\newblock Cellular neural networks: Theory.
\newblock {\em IEEE Transactions on circuits and systems}, 35(10):1257--1272,
  1988.

\bibitem{codd2014cellular}
E.~F. Codd.
\newblock {\em Cellular automata}.
\newblock Academic press, 1968.

\bibitem{conway1970game}
J.~Conway.
\newblock The game of life.
\newblock {\em Scientific American}, 223(4):4, 1970.

\bibitem{coumans2020}
E.~Coumans and Y.~Bai.
\newblock Pybullet, a python module for physics simulation for games, robotics
  and machine learning, 2016.

\bibitem{deisenroth2011pilco}
M.~Deisenroth and C.~E. Rasmussen.
\newblock Pilco: A model-based and data-efficient approach to policy search.
\newblock In {\em Proceedings of the 28th International Conference on machine
  learning (ICML-11)}, pages 465--472. Citeseer, 2011.

\bibitem{devlin2018bert}
J.~Devlin, M.-W. Chang, K.~Lee, and K.~Toutanova.
\newblock Bert: Pre-training of deep bidirectional transformers for language
  understanding.
\newblock {\em arXiv preprint arXiv:1810.04805}, 2018.

\bibitem{dosovitskiy2020image}
A.~Dosovitskiy, L.~Beyer, A.~Kolesnikov, D.~Weissenborn, X.~Zhai,
  T.~Unterthiner, M.~Dehghani, M.~Minderer, G.~Heigold, S.~Gelly, et~al.
\newblock An image is worth 16x16 words: Transformers for image recognition at
  scale.
\newblock {\em arXiv preprint arXiv:2010.11929}, 2020.

\bibitem{duan2016rl}
Y.~Duan, J.~Schulman, X.~Chen, P.~L. Bartlett, I.~Sutskever, and P.~Abbeel.
\newblock Rl2: Fast reinforcement learning via slow reinforcement learning.
\newblock {\em arXiv preprint arXiv:1611.02779}, 2016.

\bibitem{dubey2018investigating}
R.~Dubey, P.~Agrawal, D.~Pathak, T.~L. Griffiths, and A.~A. Efros.
\newblock Investigating human priors for playing video games.
\newblock {\em arXiv preprint arXiv:1802.10217}, 2018.

\bibitem{eagleman2020livewired}
D.~Eagleman.
\newblock {\em Livewired: The inside story of the ever-changing brain}.
\newblock Canongate Books, 2020.

\bibitem{esser2020taming}
P.~Esser, R.~Rombach, and B.~Ommer.
\newblock Taming transformers for high-resolution image synthesis.
\newblock {\em arXiv preprint arXiv:2012.09841}, 2020.

\bibitem{fortuin2018som}
V.~Fortuin, M.~H{\"u}ser, F.~Locatello, H.~Strathmann, and G.~R{\"a}tsch.
\newblock Som-vae: Interpretable discrete representation learning on time
  series.
\newblock {\em arXiv preprint arXiv:1806.02199}, 2018.

\bibitem{learningtopredict2019}
D.~Freeman, D.~Ha, and L.~Metz.
\newblock Learning to predict without looking ahead: World models without
  forward prediction.
\newblock In {\em Advances in Neural Information Processing Systems},
  volume~32. Curran Associates, Inc., 2019.
\newblock \url{https://learningtopredict.github.io}.

\bibitem{wann2019}
A.~Gaier and D.~Ha.
\newblock Weight agnostic neural networks.
\newblock In {\em Advances in Neural Information Processing Systems},
  volume~32. Curran Associates, Inc., 2019.
\newblock \url{https://weightagnostic.github.io}.

\bibitem{Gal2016Improving}
Y.~Gal, R.~McAllister, and C.~E. Rasmussen.
\newblock Improving {PILCO} with {B}ayesian neural network dynamics models.
\newblock In {\em Data-Efficient Machine Learning workshop, ICML}, Apr. 2016.

\bibitem{girdhar2019video}
R.~Girdhar, J.~Carreira, C.~Doersch, and A.~Zisserman.
\newblock Video action transformer network.
\newblock In {\em Proceedings of the IEEE/CVF Conference on Computer Vision and
  Pattern Recognition}, pages 244--253, 2019.

\bibitem{goyal2021recurrent}
A.~Goyal, A.~Lamb, J.~Hoffmann, S.~Sodhani, S.~Levine, Y.~Bengio, and
  B.~Sch{\"o}lkopf.
\newblock Recurrent independent mechanisms.
\newblock In {\em International Conference on Learning Representations}, 2021.

\bibitem{graves2014neural}
A.~Graves, G.~Wayne, and I.~Danihelka.
\newblock Neural turing machines.
\newblock {\em arXiv preprint arXiv:1410.5401}, 2014.

\bibitem{guttenberg2016permutation}
N.~Guttenberg, N.~Virgo, O.~Witkowski, H.~Aoki, and R.~Kanai.
\newblock Permutation-equivariant neural networks applied to dynamics
  prediction.
\newblock {\em arXiv preprint arXiv:1612.04530}, 2016.

\bibitem{ha2017evolving}
D.~Ha.
\newblock Evolving stable strategies.
\newblock {\em http://blog.otoro.net/}, 2017.

\bibitem{ha2016hypernetworks}
D.~Ha, A.~Dai, and Q.~V. Le.
\newblock Hypernetworks.
\newblock {\em arXiv preprint arXiv:1609.09106}, 2016.

\bibitem{ha2018worldmodels}
D.~Ha and J.~Schmidhuber.
\newblock Recurrent world models facilitate policy evolution.
\newblock In {\em Advances in Neural Information Processing Systems 31}, pages
  2451--2463. Curran Associates, Inc., 2018.
\newblock \url{https://worldmodels.github.io}.

\bibitem{hafner2018planet}
D.~Hafner, T.~Lillicrap, I.~Fischer, R.~Villegas, D.~Ha, H.~Lee, and
  J.~Davidson.
\newblock Learning latent dynamics for planning from pixels.
\newblock {\em arXiv preprint arXiv:1811.04551}, 2018.

\bibitem{hansen2006cma}
N.~Hansen.
\newblock The cma evolution strategy: a comparing review.
\newblock {\em Towards a new evolutionary computation}, pages 75--102, 2006.

\bibitem{haruno2001mosaic}
M.~Haruno, D.~M. Wolpert, and M.~Kawato.
\newblock Mosaic model for sensorimotor learning and control.
\newblock {\em Neural computation}, 13(10):2201--2220, 2001.

\bibitem{lstm1997}
S.~Hochreiter and J.~Schmidhuber.
\newblock Long short-term memory.
\newblock {\em Neural Computation}, 9(8):1735--1780, 1997.

\bibitem{hochreiter2001learning}
S.~Hochreiter, A.~S. Younger, and P.~R. Conwell.
\newblock Learning to learn using gradient descent.
\newblock In {\em International Conference on Artificial Neural Networks},
  pages 87--94. Springer, 2001.

\bibitem{huang2020}
W.~Huang, I.~Mordatch, and D.~Pathak.
\newblock One policy to control them all: Shared modular policies for
  agent-agnostic control.
\newblock In {\em Proceedings of the 37th International Conference on Machine
  Learning, {ICML} 2020, 13-18 July 2020, Virtual Event}, volume 119 of {\em
  Proceedings of Machine Learning Research}, pages 4455--4464. {PMLR}, 2020.

\bibitem{jaegle2021perceiver}
A.~Jaegle, F.~Gimeno, A.~Brock, A.~Zisserman, O.~Vinyals, and J.~Carreira.
\newblock Perceiver: General perception with iterative attention.
\newblock {\em arXiv preprint arXiv:2103.03206}, 2021.

\bibitem{joshi2020transformers}
C.~Joshi.
\newblock Transformers are graph neural networks.
\newblock {\em The Gradient}, 2020.

\bibitem{kirsch2020meta}
L.~Kirsch and J.~Schmidhuber.
\newblock Meta learning backpropagation and improving it.
\newblock {\em arXiv preprint arXiv:2012.14905}, 2020.

\bibitem{carracing_v0}
O.~Klimov.
\newblock Carracing-v0, 2016.

\bibitem{set2019}
J.~Lee, Y.~Lee, J.~Kim, A.~Kosiorek, S.~Choi, and Y.~W. Teh.
\newblock Set transformer: A framework for attention-based
  permutation-invariant neural networks.
\newblock In K.~Chaudhuri and R.~Salakhutdinov, editors, {\em Proceedings of
  the 36th International Conference on Machine Learning}, volume~97 of {\em
  Proceedings of Machine Learning Research}, pages 3744--3753. PMLR, 09--15 Jun
  2019.

\bibitem{lee2019network}
K.~Lee, K.~Lee, J.~Shin, and H.~Lee.
\newblock Network randomization: A simple technique for generalization in deep
  reinforcement learning.
\newblock {\em arXiv preprint arXiv:1910.05396}, 2019.

\bibitem{liu2020pic}
I.-J. Liu, R.~A. Yeh, and A.~G. Schwing.
\newblock Pic: permutation invariant critic for multi-agent deep reinforcement
  learning.
\newblock In {\em Conference on Robot Learning}, pages 590--602. PMLR, 2020.

\bibitem{luong2015effective}
M.-T. Luong, H.~Pham, and C.~D. Manning.
\newblock Effective approaches to attention-based neural machine translation.
\newblock {\em arXiv preprint arXiv:1508.04025}, 2015.

\bibitem{miconi2020backpropamine}
T.~Miconi, A.~Rawal, J.~Clune, and K.~O. Stanley.
\newblock Backpropamine: training self-modifying neural networks with
  differentiable neuromodulated plasticity.
\newblock {\em arXiv preprint arXiv:2002.10585}, 2020.

\bibitem{miconi2018differentiable}
T.~Miconi, K.~Stanley, and J.~Clune.
\newblock Differentiable plasticity: training plastic neural networks with
  backpropagation.
\newblock In {\em International Conference on Machine Learning}, pages
  3559--3568. PMLR, 2018.

\bibitem{monti2017geometric}
F.~Monti, D.~Boscaini, J.~Masci, E.~Rodola, J.~Svoboda, and M.~M. Bronstein.
\newblock Geometric deep learning on graphs and manifolds using mixture model
  cnns.
\newblock In {\em Proceedings of the IEEE conference on computer vision and
  pattern recognition}, pages 5115--5124, 2017.

\bibitem{mordvintsev2020growing}
A.~Mordvintsev, E.~Randazzo, E.~Niklasson, and M.~Levin.
\newblock Growing neural cellular automata.
\newblock {\em Distill}, 2020.
\newblock \url{https://distill.pub/2020/growing-ca}.

\bibitem{mott2019towards}
A.~Mott, D.~Zoran, M.~Chrzanowski, D.~Wierstra, and D.~J. Rezende.
\newblock Towards interpretable reinforcement learning using attention
  augmented agents.
\newblock {\em arXiv preprint arXiv:1906.02500}, 2019.

\bibitem{najarro2020meta}
E.~Najarro and S.~Risi.
\newblock Meta-learning through hebbian plasticity in random networks.
\newblock {\em arXiv preprint arXiv:2007.02686}, 2020.

\bibitem{neumann1966theory}
J.~Neumann, A.~W. Burks, et~al.
\newblock {\em Theory of self-reproducing automata}, volume 1102024.
\newblock University of Illinois press Urbana, 1966.

\bibitem{ohsawa2018neuron}
S.~Ohsawa, K.~Akuzawa, T.~Matsushima, G.~Bezerra, Y.~Iwasawa, H.~Kajino,
  S.~Takenaka, and Y.~Matsuo.
\newblock Neuron as an agent, 2018.

\bibitem{ott2020giving}
J.~Ott.
\newblock Giving up control: Neurons as reinforcement learning agents.
\newblock {\em arXiv preprint arXiv:2003.11642}, 2020.

\bibitem{randazzo2020selfclassifying}
E.~Randazzo, A.~Mordvintsev, E.~Niklasson, M.~Levin, and S.~Greydanus.
\newblock Self-classifying mnist digits.
\newblock {\em Distill}, 2020.
\newblock \url{https://distill.pub/2020/selforg/mnist}.

\bibitem{sanchezlengeling2021a}
B.~Sanchez-Lengeling, E.~Reif, A.~Pearce, and A.~Wiltschko.
\newblock A gentle introduction to graph neural networks.
\newblock {\em Distill}, 2021.
\newblock https://distill.pub/2021/gnn-intro.

\bibitem{sandler2021meta}
M.~Sandler, M.~Vladymyrov, A.~Zhmoginov, N.~Miller, A.~Jackson, T.~Madams,
  et~al.
\newblock Meta-learning bidirectional update rules.
\newblock {\em arXiv preprint arXiv:2104.04657}, 2021.

\bibitem{sandlin2019backwards}
D.~Sandlin.
\newblock The backwards brain bicycle: Un-doing understanding, 2019.

\bibitem{schmidhuber1992learning}
J.~Schmidhuber.
\newblock Learning to control fast-weight memories: An alternative to dynamic
  recurrent networks.
\newblock {\em Neural Computation}, 4(1):131--139, 1992.

\bibitem{schmidhuber1993reducing}
J.~Schmidhuber.
\newblock Reducing the ratio between learning complexity and number of time
  varying variables in fully recurrent nets.
\newblock In {\em International Conference on Artificial Neural Networks},
  pages 460--463. Springer, 1993.

\bibitem{schmidhuber1993self}
J.~Schmidhuber.
\newblock A ‘self-referential’weight matrix.
\newblock In {\em International Conference on Artificial Neural Networks},
  pages 446--450. Springer, 1993.

\bibitem{sorokin2015deep}
I.~Sorokin, A.~Seleznev, M.~Pavlov, A.~Fedorov, and A.~Ignateva.
\newblock Deep attention recurrent q-network.
\newblock {\em arXiv preprint arXiv:1512.01693}, 2015.

\bibitem{strubell2019energy}
E.~Strubell, A.~Ganesh, and A.~McCallum.
\newblock Energy and policy considerations for deep learning in nlp.
\newblock {\em arXiv preprint arXiv:1906.02243}, 2019.

\bibitem{sudhakaran2021growing}
S.~Sudhakaran, D.~Grbic, S.~Li, A.~Katona, E.~Najarro, C.~Glanois, and S.~Risi.
\newblock Growing 3d artefacts and functional machines with neural cellular
  automata.
\newblock {\em arXiv preprint arXiv:2103.08737}, 2021.

\bibitem{sun2019learning}
C.~Sun, F.~Baradel, K.~Murphy, and C.~Schmid.
\newblock Learning video representations using contrastive bidirectional
  transformer.
\newblock {\em arXiv preprint arXiv:1906.05743}, 2019.

\bibitem{attentionagent2020}
Y.~Tang, D.~Nguyen, and D.~Ha.
\newblock Neuroevolution of self-interpretable agents.
\newblock In {\em Proceedings of the Genetic and Evolutionary Computation
  Conference}, 2020.
\newblock \url{https://attentionagent.github.io}.

\bibitem{van2008visualizing}
L.~Van~der Maaten and G.~Hinton.
\newblock Visualizing data using t-sne.
\newblock {\em Journal of machine learning research}, 9(11), 2008.

\bibitem{vaswani2017}
A.~Vaswani, N.~Shazeer, N.~Parmar, J.~Uszkoreit, L.~Jones, A.~N. Gomez, L.~u.
  Kaiser, and I.~Polosukhin.
\newblock Attention is all you need.
\newblock In I.~Guyon, U.~V. Luxburg, S.~Bengio, H.~Wallach, R.~Fergus,
  S.~Vishwanathan, and R.~Garnett, editors, {\em Advances in Neural Information
  Processing Systems}, volume~30. Curran Associates, Inc., 2017.

\bibitem{velivckovic2017graph}
P.~Veli{\v{c}}kovi{\'c}, G.~Cucurull, A.~Casanova, A.~Romero, P.~Lio, and
  Y.~Bengio.
\newblock Graph attention networks.
\newblock {\em arXiv preprint arXiv:1710.10903}, 2017.

\bibitem{wang2016learning}
J.~X. Wang, Z.~Kurth-Nelson, D.~Tirumala, H.~Soyer, J.~Z. Leibo, R.~Munos,
  C.~Blundell, D.~Kumaran, and M.~Botvinick.
\newblock Learning to reinforcement learn.
\newblock {\em arXiv preprint arXiv:1611.05763}, 2016.

\bibitem{wang2020linformer}
S.~Wang, B.~Li, M.~Khabsa, H.~Fang, and H.~Ma.
\newblock Linformer: Self-attention with linear complexity.
\newblock {\em arXiv preprint arXiv:2006.04768}, 2020.

\bibitem{wolfram1984cellular}
S.~Wolfram.
\newblock Cellular automata as models of complexity.
\newblock {\em Nature}, 311(5985):419--424, 1984.

\bibitem{wu2020comprehensive}
Z.~Wu, S.~Pan, F.~Chen, G.~Long, C.~Zhang, and S.~Y. Philip.
\newblock A comprehensive survey on graph neural networks.
\newblock {\em IEEE transactions on neural networks and learning systems},
  2020.

\bibitem{xiong2021nystr}
Y.~Xiong, Z.~Zeng, R.~Chakraborty, M.~Tan, G.~Fung, Y.~Li, and V.~Singh.
\newblock Nystr{\"o}mformer: A nystr{\"o}m-based algorithm for approximating
  self-attention.
\newblock {\em arXiv preprint arXiv:2102.03902}, 2021.

\bibitem{yun2019graph}
S.~Yun, M.~Jeong, R.~Kim, J.~Kang, and H.~J. Kim.
\newblock Graph transformer networks.
\newblock {\em arXiv preprint arXiv:1911.06455}, 2019.

\bibitem{zaheer2017deep}
M.~Zaheer, S.~Kottur, S.~Ravanbakhsh, B.~Poczos, R.~Salakhutdinov, and
  A.~Smola.
\newblock Deep sets.
\newblock {\em arXiv preprint arXiv:1703.06114}, 2017.

\bibitem{zambaldi2018deep}
V.~Zambaldi, D.~Raposo, A.~Santoro, V.~Bapst, Y.~Li, I.~Babuschkin, K.~Tuyls,
  D.~Reichert, T.~Lillicrap, E.~Lockhart, M.~Shanahan, V.~Langston, R.~Pascanu,
  M.~Botvinick, O.~Vinyals, and P.~Battaglia.
\newblock Deep reinforcement learning with relational inductive biases.
\newblock In {\em International Conference on Learning Representations}, 2019.

\bibitem{zhang2021learning}
D.~Zhang, C.~Choi, J.~Kim, and Y.~M. Kim.
\newblock Learning to generate 3d shapes with generative cellular automata.
\newblock {\em arXiv preprint arXiv:2103.04130}, 2021.

\bibitem{zhang2018gaan}
J.~Zhang, X.~Shi, J.~Xie, H.~Ma, I.~King, and D.-Y. Yeung.
\newblock Gaan: Gated attention networks for learning on large and
  spatiotemporal graphs.
\newblock {\em arXiv preprint arXiv:1803.07294}, 2018.

\bibitem{deepPILCOgithub}
X.~Zuo.
\newblock Pytorch implementation of improving pilco with bayesian neural
  network dynamics models, 2018.
\newblock \url{https://github.com/zuoxingdong/DeepPILCO}.

\end{thebibliography}
\bibliographystyle{abbrv}

\end{document}